\documentclass[conference]{IEEEtran}
\usepackage{amsmath,amssymb,amsfonts}
\usepackage[normalem]{ulem}
\usepackage{algorithm}
\usepackage[noend]{algpseudocode}
\usepackage{graphicx}
\usepackage{textcomp}
\usepackage{pgfplots}
\pgfplotsset{compat=1.16}
\usepgfplotslibrary{statistics}
\usepackage{xcolor}
\usepackage{epsfig,color,ulem,url,booktabs}
\usepackage{multicol}
\usepackage{multirow}
\newsavebox{\tablebox}
\usepackage{breqn}

\usepackage{subfigure}
\usepackage{comment}
\usepackage{ulem}

\IEEEoverridecommandlockouts    
\makeatletter
\algnewcommand{\LineComment}[1]{\Statex \hskip\ALG@thistlm \(\triangleright\) #1}
\makeatother
\newcommand\defeq{\mathrel{\stackrel{\makebox[0pt]{\mbox{\normalfont\scriptsize def}}}{:=}}}

\newcommand{\sig}[1]{{\small\textsf{{#1}}}}

\begin{document}

\title{\LARGE \bf Potential-based Credit Assignment for Cooperative RL-based Testing of Autonomous Vehicles}

\author{Utku Ayvaz$^{1,2}$, Chih-Hong Cheng$^{1}$,  and Shen Hao$^{2,3}$%
\thanks{$^{1}$ Fraunhofer IKS, Germany}
\thanks{$^{2}$ Technical University of Munich, Germany}
\thanks{$^{3}$ fortiss GmbH, Germany} 
\thanks{This project is funded by the StMWi Bayern as part of a project to support the thematic development of the Fraunhofer IKS. Correspondence to \{uktu.ayvaz,chih-hong.cheng\}@iks.fraunhofer.de,  hao.shen@tum.de}
}

\maketitle

\begin{abstract}

While autonomous vehicles (AVs) may perform remarkably well in generic real-life cases, their irrational action in some unforeseen cases leads to critical safety concerns. This paper introduces the concept of collaborative reinforcement learning (RL) to generate challenging test cases for AV planning and decision-making module. One of the critical challenges for collaborative RL is the credit assignment problem, where a proper assignment of rewards to multiple agents interacting in the traffic scenario, considering all parameters and timing, turns out to be non-trivial. In order to address this challenge, we propose a novel potential-based reward-shaping approach inspired by counterfactual analysis for solving the credit-assignment problem. The evaluation in a simulated environment demonstrates the superiority of our proposed approach against other methods using local and global rewards.

\end{abstract}


\section{Introduction}

Ensuring the safety of Autonomous Vehicles (AV) with efficiency has been a dominating factor in closing the gap from running demonstrations to safe products. Among various dimensions to be considered, such as perception robustness or operational safety, one crucial aspect is verifying and validating the AV stack. Apart from scenario replays or simple random testing, intelligent testing techniques utilizing classical software testing paradigm and control theory have been currently under active research (to name a few recent results~\cite{piazzoni2021vista,metzger2021towards,ebadi2021efficient,kaufmann2021critical,li2022comopt}), where fundamental techniques such as equivalence class testing or optimization-based testing are being extended by the research community to allow the efficient finding of scenarios that challenges the AV.  
  
In this paper, inspired by the success of using reinforcement learning (RL) in games~\cite{mnih2013playing,silver2016mastering,perolat2022mastering}, we study how RL can be used in generating test case scenarios. In games such as chess or the game of GO, the controller learned by RL interacts with the opponent interactively, where based on the current system state and the move of the opponent, provides the next move. For the testing of AV, one can analogously view each agent in the road traffic (referred to as non-player-character NPC) to be a chess piece controlled by RL, where the undesired situation for the AV can be assigned as the reward of RL. This altogether establishes the basic framework. 

Nevertheless, in realizing the test case generation method using RL, several non-trivialities arise.  Apart from setting the undesired situation (e.g., motion jerks or collision) into concrete numerical rewards, one aspect is related to the enormous  action space caused by the product of actions for individual agents, making an efficient convergence of good testing strategy infeasible. 
Thus, we propose using Multi-Agent Reinforcement Learning (MARL)~\cite{bucsoniu2010multi} beyond simple RL for training agents. Each agent in MARL is trained with a projection operator separating its own state and the environment state (for other agents and AV), and the reward is assigned accordingly.   The design of reward assignment thus changes the agent's behavior, wherein in a fully cooperative setup, an agent shares its reward with the other agents as a common reward. However, due to the use of a common reward, each agent is being obfuscated on the value of its individual action. To mediate this issue, we further propose a novel Potential-based Reward Shaping (PBRS) function that intuitively only rewards an agent if it contributes to the reward of other agents via a potential function inspired by counterfactual analysis. We also create a framework for efficient training of independent, collaboratively behaved agents. 

For evaluation, we created a simulation-based environment and evaluated the performance of MARL-based testing techniques against different collaborative schemes. In contrast to standard competitive or naïve collaborative approaches, our PBRS-based credit assignment approach leads to considerably superior performance in generating crash scenarios. 

The rest of the paper is structured as follows. After Section~\ref{section:related} highlighting related work, we in Section~\ref{section:background} lay the foundation for RL-based testing for AV.  In Section~\ref{section:PBRS} we present the key results, namely the PBRS and the agent training scheme. Finally, we summarize our evaluation in Section~\ref{sec:evaluation} and conclude with future work in Section~\ref{sec:concluding.remarks}.

\section{Related Work}\label{section:related}

Intelligent testing of autonomous driving has been under active research. It is also considered important by many AV safety standards (e.g., ISO TS 5083) 
under development, as testing methods beyond scenario replay are crucial in uncovering previously unseen situations. The research trend can be observed in the recent IEEE competition on Autonomous Driving Testing, where the open-source Baidu Apollo version~6.0 is chosen as the system under test~\cite{lemke2021}. For intelligent testing, the work by Ebadi et al.~\cite{ebadi2021efficient} applied genetic programming to search for crash scenarios, where the fitness function characterizes the number of accidents being generated (the more, the better) as well as the distance being traveled (the shorter, the better), where the control of the action sequence is on the signal level. Another recent work by Kaufmann et al.~\cite{kaufmann2021critical} also applied genetic algorithms where the action for each agent is on the semantic level, such as moving forward and cutting left. The joint result of applying VerifAI and SCENIC~\cite{viswanadha2021addressing} applied multiple optimization algorithms such as Bayesian optimization. Seymour, et al.~\cite{seymour2021empirical} applied the concept of equivalence class partitioning and metamorphic testing for scenario generation. The SALVO approach~\cite{nguyen2021salvo} approach uses a combination of equivalence class testing and random testing, where it developed pair-wise metrics to capture the diversity between trajectories, thereby disallowing similar trajectories to be re-sampled and re-evaluated. Finally, the ComOpT approach~\cite{li2022comopt} uses a combination of equivalence class and random testing to find diversified scenarios while, with a given scenario, generating a perturbed scenario via local search.  Our method of using RL-based testing complements these techniques in that it is commonly believed in the industry that there is no single silver bullet, implying that practically one needs a portfolio of testing methods. Methodologically, our proposed RL-based approach is also an optimization-based approach directed towards bug finding (in contrast to undirected approaches such as random testing), where the learning of such agents is made efficient by the use of MARL.

We use MARL as our fundamental  engineering paradigm for test case generation, where in this paper we detail the underlying technical challenge on credit assignment. Within credit assignment, the work of Devlin et al.~\cite{devlin2014potential} combines difference rewards and potential-based reward shaping to shape the reward of independent RL agents to capture the contribution of each agent to systems performance. Another approach is the use of COMA~\cite{foerster2018counterfactual}, where a centralized critic model is trained to learn the Q-function of the centralized system. Then this critic model is used to marginalize the action of an agent by fixing the actions of other agents.  Another work~\cite{feng2022multi} proposes performing credit assignments on multi-levels, where they propose a hierarchical model that is divided into manager and worker levels. Then at each level, a centralized critic is designed to learn the credit-assignment scheme of that level. Finally, in the recent work of Xiao, Ramasubramanian, and Poovendran~\cite{xiao2021shaping}, potential-based reward shaping is proposed again to solve the credit-assignment problem, where the authors proposed shaping advice in deep MARL to shape the reward function with expert knowledge. The common ground we have with previous methods is that we also shape the reward of each agent in the system and use potential-based reward shaping. However, the critical  difference is that we propose  a potential function that rewards agents solely according to their collaboration with other agents. Our designed potential function is thus not application specific.

\section{Reinforcement Learning for AV Testing}\label{section:background}
In this section, we briefly introduce the concepts of MARL and its applications in AV testing.

\subsection{Foundation}
Reinforcement learning (RL) is an area of machine learning concerned with how intelligent agents ought to take actions in an environment to maximize the notion of cumulative reward. Within the model-free setup, Q-learning~\cite{watkins1989learning}, a classic RL algorithm, performs an update on the quality of the state-action pair  $Q(s,a)$ via the following equation, i.e.,
\begin{equation*}
    Q(s,a) \gets Q(s,a)+ \beta(r(s, a)+ \alpha \max_{a'} [Q(s',a')-Q(s,a)]),
\end{equation*}
where $s$ is the state, $a$ is the action of the agent, the $r(s,a,s')$ is the reward signal of the agent, $\beta$ is the learning rate, $\alpha$ is the discount factor, $s'$ and $a'$ are future state and action, respectively. 

In the testing of AV, we aim at training a controller (i.e., a reactive test case generator) to steer all movements of all non-player characters (NPCs), including vehicles, pedestrians, or traffic lights, i.e., all agents beyond the autonomous vehicles. The action space thus corresponds to all possible movements of all NPCs, and the state space corresponds to observable states from the controller. However, some of the AV's internal states may not be accessible to the controller. The reward can be translated from undesired behaviors such as jerky motions, near hits or collisions. After the controller signaling actions for each NPC, the time elapses for a small amount~$\Delta$ (e.g., $0.1$~sec), and the state is updated accordingly (in a simulated environment). 

When multiple agents need to be controlled by a single controller, overall one has the joint 
state space $S = S_0 \times S_1 \times S_2 \times \ldots \times S_n$ and the joint action space $A = A_1 \times A_2 \times \ldots \times A_n$, where $S_k$, $A_k$ is the  state and action spaces of agent $k$, respectively. We reserve index~$0$ to represent the observable state of the AV; in the action space we do not have $A_0$, as the action of the AV is not controllable by the test case generator.  Given $s\in S$ and action $a \in A$, the system receives the global reward $r(s, a)$. 

It is well known from the literature~\cite{busoniu2008comprehensive} that training such a controller is difficult due to the product of the state and action space of agents being prohibitively large, leading to methods such as MARL to enable computationally efficient training.

\subsection{Multi-Agent Reinforcement Learning (MARL), Laziness and Credit Assignment}

MARL proposes training each agent in a decentralized manner, where each agent has its own state and action space. To train these agents, there are two types of reward signals. The first reward signal is the \emph{local reward}, where each agent receives only a reward based on their state and actions. For agent~$i$, let $r_i(s, a_i)$ be the corresponding local reward. In a \emph{fully cooperative setup}, one trivial way of setting the local reward is to allow local reward to be the same as the global reward, i.e., for $s \in S$ and $a\in A$, $r_i(s, a_i):=r(s, a)$. Nevertheless, as each agent receives the same global reward, this also allows bad-behaving agents to receive positive rewards due to the good behavior of other agents. This phenomenon is known as the \emph{laziness}~\cite{kravarisdeep}, and it significantly hinders the generation of well-performed agents. In order to avoid this phenomenon, fair arbitration should be performed such that each agent receives a reward according to their contribution to the global reward. The arbitration is referred to as the \emph{credit assignment problem}.

\subsection{MARL for AV Testing}

To use MARL for AV testing purposes, we detail how the local reward $r_i(s, a_i)$ and the global reward $r(s, a)$ are computed. In this paper, we set a simple objective for each agent (NPC) to make the AV act irrationally. Therefore, we reward an agent when the front part of the AV crashes with the agent. We also set an agent to receive a negative reward if it crashes with another agent that is not the AV. Note that the presented method is generic; one may also use other definitions for global and local rewards. 

Precisely, for every agent $i$, we define its interaction with other agent (indexed $j$) and the AV (indexed $0$) as follows. 
\begin{equation}\label{kappa}
    crash(s_i,s_j) \defeq
    \begin{cases}
    \kappa_1, & \text{if $j=0$ and front-crash occurs}\\
    -\kappa_2, & \text{if $j\neq 0$ and crash occurs} \\
    0 & \text{otherwise} \\
\end{cases}
\end{equation}

The local reward for agent~$i$ is defined by considering the interaction with other agents, while the global reward is defined as the sum of all local rewards.

\begin{equation}\label{eq.competitive.reward}
r_i(s, a_i) \defeq \sum^{n}_{j=0, j\neq i} crash(s_i, s_j)
\end{equation}

\begin{equation}\label{eq.naive.cooperative.reward}
r(s, a) \defeq  \sum^{n}_{j=1} r_i(s, a_i)
\end{equation}

To implement a fully cooperative MARL, one may replace $r_i(s, a_i)$ and use~$r(s, a)$ as the local reward for an agent, implying that the goal of training an individual agent is to maximize the global good. A fully competitive MARL, on the other hand, simply retains using~$r_i(s,a_i)$ as the local reward. It leads to competitive behavior as each agent tries to maximize its good. As depicted in later sections, \emph{reward shaping} is a method that works in between the naïve collaborative and the fully competitive reward assignment strategy.

\subsection{Potential-based Reward Shaping}

The potential-based reward shaping~\cite{ng1999policy} technique creates a modified reward $r^{PB}(s) $ for a cooperative MARL agent~$i$ via the following definition.

\begin{equation}\label{eq.standard.reward.shaping}
    r^{PB}(s,a) \defeq r(s,a)+F(s,s')
\end{equation}

In the definition, $s'\in S$ is the successor state of moving from state~$s$ with action~$a$. $F(s, s')$ is the potential function $F(s,s') = \Phi(s)-\alpha \;\Phi(s')$ with $\Phi: S\rightarrow \mathbb{R}$ mapping from any state in $S$ to a real value. In the following section, we detail how we design such $\Phi(s,a)$ to be task-independent.

\section{Task Independent Potential-based Reward Shaping via Counterfactual Analysis}\label{section:PBRS}

\subsection{Designing Potential Functions inspired by Counterfactual Analysis}

We now present our potential-based reward-shaping technique inspired by \emph{counterfactual analysis}. The underlying idea is the following: if the existence of the agent at time $t$ does not change the overall positive reward, intuitively, we know that it should not take a share of the global reward. As an example, a car being very far from the location of an accident, even when one removes it, may not change the outcome of the accident. Therefore, the action of the car should not be credited.

We introduce some notations to allow detailing the counterfactual analysis. Let $f$ be the environment update function that changes state~$s^t$ (state at time  $t$) to $s^{t+1}$ (the state at time $t+1$) with agents using action $a$ at time $t$ (denoted as $a^t$), i.e.,  $s^{t+1} \defeq f(s^t,a^t)$. Let $s_{-i}^{t+1} \defeq f_{-i}(s_{-i}^t, a_{-i}^t)$ be the corresponding state update without agent~$i$ being present, where use $s^t_{-i}$ and $a^t_{-i}$ to denote the state and action at time~$t$ without agent~$i$ and $f_{-i}$ as the update function where agent~$i$ does not exist. 

With the above notation, we now explain the intuition behind the design of the potential function. 

\begin{itemize}
    \item  We perform a counterfactual analysis by considering the impact of agent~$i$ being absent from time~$t$ onward. We use $s^{t+1,c(t)}$ to represent the counterfactual state (we use $c(t)$ to abbreviate ``counterfactual at time~$t$"), leading to the following definition, with $\pi_{-i}$ being the policy of other agents. 

\vspace{-5mm}

\begin{equation}
    s^{t+1,c(t)} \defeq f_{-i}(s^t_{-i},\pi_{-i}(s^t_{-i}))
\end{equation}

Continuing the counterfactual analysis, then we have the following state update with $k\in \mathbf{N}^+$ under the assumption that agent $i$ no longer disappears from time $t$ onward.
\begin{equation}
    s^{t+k+1,c(t)} \defeq f_{-i}(s^{t+k,c(t)},\pi_{-i}(s^{t+k,c(t)}))
\end{equation} 

\item As the reward at state $s^{t+k+1,c(t)}$ equals $r_{-i}(s^{t+k+1,c(t)})$, we define $r^{c(t)}_{-i,acc}$, the accumulated discounted reward from $t$ onward (i.e., in Formula~\eqref{eq.t.counterfactual}, start the index with $k=0$), provided that  counterfactual analysis occurs at time~$t$ without agent~$i$.

\vspace{-3mm}
\begin{dmath}\label{eq.t.counterfactual}
    r^{c(t)}_{-i,acc}(s^t)  \defeq \sum_{k=0}^{\infty} \alpha^{k} r_{-i}(s^{t+1+k,c(t)})
    =  r_{-i}(s^{t+1, c(t)}) + \alpha\; r_{-i}(s_t^{t+2, c(t)}) + \alpha^2 \; r_{-i}(s^{t+3, c(t)}) + \ldots   
\end{dmath}

\item  We now consider the alternative case where agent~$i$ disappears from time~$t+1$ onward. This means that the system first evolves from state $s^{t}$ to $s^{t+1}$ with agent~$i$ being present. 
Subsequently, as agent~$i$ counterfactually disappears from time~$t+1$ onward, we compute the next states as well as the associated accumulated reward~$r^{c(t+1)}_{-i,acc}$ also from time $t+1$ onward (i.e., in Formula~\eqref{eq.t_plus_1.counterfactual}, start the index with $k=1$).

\vspace{-3mm}
\begin{dmath}\label{eq.t_plus_1.counterfactual}
     r^{c(t+1)}_{-i,acc}(s^t)  \defeq \sum_{k=1}^{\infty} \alpha^{k} r_{-i}(s^{t+1+k, c(t+1)}) = \alpha\;r_{-i}(s^{t+2, c(t+1)}) + \alpha^2 \;r_{-i}(s^{t+3, c(t+1)})  + \alpha^3 \;r_{-i}(s^{t+4, c(t+1)}) +\ldots
\end{dmath}

\end{itemize}

We define the potential function  $\Phi(s^t)$ as follows. 

\vspace{-3mm}
\begin{dmath}\label{eq.potential}
    \Phi_c(s^t) \defeq -(r_{-i}(s^{t+1,c(t)})+\alpha\; r_{-i}(s^{t+2,c(t)})+\alpha^2\; r_{-i}(s^{t+3,c(t)}) + \ldots)
\end{dmath}

With $F(s^t, s^{t+1}) \defeq \Phi_c(s^t) - \alpha \Phi_c(s^{t+1})$, it is a simple exercise to derive that $\Phi_c(s^t) - \alpha \Phi_c(s^{t+1}) = r^{c(t+1)}_{-i,acc}(s^t)  - r^{c(t)}_{-i,acc}(s^t)$, where by reorganizing the terms, one achieves the following equation. 

\vspace{-2mm}

\begin{dmath}\label{eq:rewriting}
r^{c(t+1)}_{-i,acc}(s^t)  - r^{c(t)}_{-i,acc}(s^t)  \\= -r_{-i}(s^{t+1,c(t)}) \\ - \alpha\;(r_{-i}(s^{t+2, c(t)}) - r_{-i}(s^{t+2, c(t+1)})) - \alpha^2(r_{-i}(s^{t+3, c(t)}) - r_{-i}(s^{t+3, c(t+1)}))  - \ldots
\end{dmath}

Finally, for numerical reasons, in contrast to Formula~\eqref{eq.standard.reward.shaping} we add a scaling value $\gamma$ to the Formula~\eqref{eqn:PBRS} to scale how much to punish or reward the agents for their interactions with other agents. As such, our counterfactual-driven potential-based reward shaping function equals the following:
\begin{equation}
    \label{eqn:PBRS}
     r^{PB}(s^t, a) \defeq r(s^t, a)+\gamma(\Phi_c(s^t)-\alpha \Phi_c(s^{t+1}))
\end{equation}

With the above formulation, we now offer an intuition on our reward shaping technique: In Formula~\eqref{eq:rewriting}, the first term $r_{-i}(s^{t+1,c(t)})$ is the immediate reward of other agents at $t+1$ when agent $i$ is absent, where the negative sign in the front highlights that the term should be subtracted. This matches the intuition where if agent~$i$ at time~$t$ does not actively contribute to the created accident, then $r_{-i}(s^{t+1,c(t)})$ should be large (i.e., other agents are actively contributing), thereby implying that the reward for agent~$i$ at time~$t$ should be small. For subsequent terms, they are the reward difference on counterfactual analysis being discounted.

\subsection{Learning}

\begin{algorithm*}
    \caption{MARL with counterfactual analysis inspired reward shaping (homogeneous agent)}\label{algo:Framework}
    \textbf{Input:} $\alpha$ discount factor, $\gamma$ hyper-parameter for PBRS, $\sig{uim}$, $\sig{uic}$, $\sig{maxtimesteps}$, $\sig{batch-size}$
    
    \textbf{Output:} $Q^{\pi}$ is the policy function approximator DNN that takes observation from an agent to be controlled (i.e., NPC) and returns the vector of expected discounted reward for each possible action.
    \begin{algorithmic}[1]

  \State Let $Q^{\phi}$ be another DNN that takes the same input and output as $Q^{\pi}$. 
    \State Initialize the weights for $Q^{\pi}$ and $Q^{\phi}$ \label{networkinit}
    \State Initialize the memory $B$ \Comment{Used to store the memories of the simulations for Q-learning}
    \State $Q^{\pi}_{target}\gets Q^{\pi}$; $Q^{\phi}_{target}\gets Q^{\phi}$  \Comment{$Q^{\pi}_{target}$ and $Q^{\phi}_{target}$ are updated periodically} \label{maineq} 

    \State \sig{timestep} = 0
        \While{$\sig{timestep} < \sig{maxtimesteps}$} 
        
            \State \sig{episode-finish} $\gets$ \sig{false}\label{Episodeinit} 
           \State Randomly initialize an initial system state $s=(s_1,s_2,\ldots,s_n) \in S$.\label{initstates} 
 \While{\sig{episode-finish} = \sig{false}}
                \State $a \gets (\sig{argmax}Q^{\pi}(\sig{obs}_1(s)), \ldots, \sig{argmax} Q^{\pi}(\sig{obs}_n(s)))$ \Comment{Output of actions for all agents decided by $Q^{\pi}$}\label{getactions}
                \LineComment{Perform a simulation step to derive the next state and information whether the agent has achieved its end state}
                
                \State $s_{next}, r(s, a) \gets \sig{simulation-step}(s, a)$

                \LineComment{Counterfactual agent removal on state $s$ to get an estimate on $\Phi(s)$}

                \State $s^{c} \gets  \sig{counterfactual-agent-removal}(s)$  
             \State Take an agent indexed $j_1$ that is still active after counterfactual agent removal from $s^{c}$.
            \State $a^{c}_{j_1} \gets \sig{argmax}\; Q^\phi_{target}(\sig{obs}_{j_1}(s^{c}))$

             \State $\Phi(s) \gets - \langle Q^{\phi}_{target}(\sig{obs}_{j_1}(s^{c})) \rangle_{a^{c}_{j_1}} $ \Comment{Take the $a^{c}_{j_1}$-th output from $Q^{\phi}_{target}$, for agent $j_1$}     

  \LineComment{Counterfactual simulation to get the immediate reward}
            \State $a^c \gets (\sig{argmax}Q^{\pi}(\sig{obs}_1(s^c)), \ldots, \sig{argmax} Q^{\pi}(\sig{obs}_n(s^c)))$ \Comment{Output of actions for all agents decided by $Q^{\pi}$}\label{getactions}          
 \State $s^{c}_{succ}, r_{-i}(s^c, a^c) \gets \sig{simulation-step}(s^c, a^c)$

            \LineComment{Counterfactual agent removal on state $s_{next}$ to get an estimate on $\Phi(s_{next})$}
                \State $s^{c}_{next} \gets  \sig{counterfactual-agent-removal}(s_{next})$ 

             \State Take an agent indexed $j_2$ that is still active after counterfactual simulation from $s^{c}_{next}$.

\State $a^{c}_{next,j_2} \gets \sig{argmax}\; Q^\phi_{target}(\sig{obs}_{j_2}(s^{c}_{next}))$

    \State $\Phi(s_{next}) \gets - \langle Q^{\phi}_{target}(\sig{obs}_{j_2}(s^{c}_{next}) \rangle_{a^{c}_{next,j_2}} $ \Comment{Take the $a^{c}_{next,j_2}$-th output from $Q^{\phi}_{target}$, for agent $j_2$}

                \State $coop\_dif \gets \gamma(\Phi(s)-\alpha \Phi(s_{next}))$\Comment{Compute $F(s,s_{next})$}\label{coopdifference}

                \State $B \gets B \cup \{(s, s_{next}, s^c, s^c_{succ}, a^c, coop\_dif, r(s,a), r_{-i}(s^c, a^c), a)\}$ \Comment{Add the information to the buffer}\label{adddata}
                \If{all agents at $s_{next}$ can not move further due to end of simulation steps or due to vehicle collision}
                    \State \sig{episode-finish} $\gets$ \sig{true}
                \Else
                \State $s\gets s_{next};\; \sig{timestep} \gets \sig{timestep} + 1$
                
                \EndIf

            \EndWhile\label{euclidendwhile}
                           
             \State Train $Q^{\pi}$ to minimize the MSE loss by taking \sig{batch-size} of samples from $B$, where for each sample, its label is $r(s,a) + coop\_dif +\alpha \max_A Q^{\pi}_{target}(\sig{obs}_j(s_{next}))$ and the prediction is $\langle Q^{\pi}(\sig{obs}_j(s)) \rangle_{a_j}$ (when~$s_{next}$ is a terminal state, use $r(s,a) + coop\_dif$  as the label).

            \State Train $Q^{\phi}$ to minimize the MSE loss by taking \sig{batch-size} of samples from $B$, where  for each sample, its label is $r_{-i}(s^c,a^c) +\alpha \max_A Q^{\phi}_{target}(\sig{obs}_j(s^c_{succ}))$ and the prediction is $Q^{\phi}(\sig{obs}_j(s^c))\rangle_{a^c_j}$ (when~$s^c_{succ}$ is a terminal state, use $r_{-i}(s^c,a^c)$ as the label). 
          
                \If{$\sig{timestep} \% \sig{uim}=0$}
\State $Q^{\pi}_{target} \gets Q^{\pi}$
                \EndIf

                \If{$\sig{timestep} \% \sig{uic}=0$}
           \State    $Q^{\phi}_{target} \gets Q^{\phi}$
           \EndIf
        \EndWhile
        \State return $Q^{\pi}$
     \end{algorithmic}
     \label{alg:compress}

    \end{algorithm*}

For our reward shaping approach to be actionable, the key is to implement the counterfactual analysis as governed by Formulae~\eqref{eq.t.counterfactual} and~\eqref{eq.t_plus_1.counterfactual}. This turns out to be challenging, as one requires implementing counterfactual analysis until all agents (NPCs in the test scenario) reach their terminal state. Our proposal is to learn a helper Q-function $Q^{\phi}$ that approximates the right-hand part $(r_{-i}(s^{t+1,c(t)})+\alpha\; r_{-i}(s^{t+2,c(t)})+\alpha^2\; r_{-i}(s^{t+3,c(t)}) + \ldots)$ of Formula~\ref{eq.potential} 
by only looking at the counterfactual state $s^{t,c(t)}$. We also train another network $Q^{\pi}$ that produces the policy of an agent, which corresponds to the action of the associated NPC to be taken in the test case. 

Algorithm~\ref{algo:Framework} provides technical details on how we perform the agent training, with the simplification that all agents are homogeneous. At line~4, we first create two copies (line 3, 4)  for $Q^{\phi}$ and $Q^{\pi}$ that only performs the update in a periodic fashion via parameter $\sig{uim}$ and $\sig{uic}$ (lines~30 to~33).

In line~8, the algorithm starts by randomly initializing a state $s=(s_1,\ldots,s_n)$ to play the episode until finish (line~9). During the play, the algorithm in line~10 uses the learned policy $Q^{\pi}$ to create actions for each agent, by translating~$s$ in the simulation into the local observable state via function $\sig{obs}_i$ (the standard utility provided by the simulation engine). Our algorithmic simplification on homogeneous agents implies that the same $Q^{\pi}$ can be used for controlling all agents. With the action of each agent (NPC) made available, perform a simulation step via $\sig{simulation-step}(s,a)$ to obtain the next state $s_{next}$ and the collaborative reward $r(s,a)$ defined in Formula~\ref{eq.naive.cooperative.reward}. 

Subsequently, lines~12 to~15 perform a counterfactual analysis on state~$s$ to get an estimate on $\Phi(s)$. First, \sig{counterfactual-agent-removal} picks an agent that can still move and let it counterfactually disappear, where in the counterfactual state~$s^c$, the local state value turns to be a particular numerical value reflecting the invalidity.  Where as our $Q^{\phi}_{target}$  also takes the same input as $Q^{\pi}$, we pick one arbitrary agent $j_1$, decides its action $a^c_{j_1}$ and use the $a^c_{j_1}$-th output of $Q^{\phi}$ to be the estimate. 
To know the real reward of the counterfactual state that allows training $Q^{\phi}$, line~16 computes the counterfactual action and line~17 performs an additional simulation. 
Similarly, lines~18 to~22 perform similar actions but on the next state~$s_{next}$. In our implementation, one executes lines~12 to~22 multiple times to get a better unbiased estimate on $\Phi(s)$ and $\Phi(s_{next})$, but for simplicity, this part is omitted in Algorithm~\ref{algo:Framework}. Finally, the information is stored in the memory (line~23) in order to be used for improving the DNNs. Lines~24 to~27 perform a check if further state update (i.e., update the current state $s$ with $s_{next}$) is needed. Finally, lines~28 and~29 perform standard Q-training of $Q^{\pi}$ and $Q^{\phi}$ by minimizing the mean square errors (MSE). The label for $Q^{\pi}$ is estimated using the Bellman equation. For $Q^{\phi}$, the creation of the label also takes the recursive form as stated in Formula~\ref{eq.potential}.

\vspace{-2mm}
\section{Experimental Evaluation}\label{sec:evaluation}

To understand the effectiveness of the developed PBRS framework, we conducted an experiment by modifying the AI Gym-like driving minimalistic simulator~\cite{highway-env}, where we created multiple maps beyond simple intersections. Our testing framework creates agents as other road vehicles, where each vehicle has a discrete action space of \sig{acceleration},  \sig{deceleration}, \sig{steer-to-left}, \sig{steer-to-right}, and \sig{keep-steering}. The states of each vehicle, such as headings and velocities, are reflected as the information using a birds-eye view (BEV) under a global coordinate system. The relative observation for a particular agent is done by coordinate transformation processed in the simulation environment. 

Before applying our MARL setup to create specialized agents (NPCs) that lead to collisions, we first need to create an autonomous vehicle that can be tested against. For this, we also train another AV that operates on a standard environment with some normally operating NPCs~\cite{doi:10.3141/1999-10,treiber2000congested}. 
We then apply the MARL framework with three reward assignment strategies, namely, 
\begin{enumerate}
    \item the fully competitive setup where each agent takes the reward as governed by Formula~\eqref{eq.competitive.reward},
    \item  the naïve collaborative setup by using Formula~\eqref{eq.naive.cooperative.reward} as the local reward for each agent, and finally
    \item the counterfactual-driven potential-based reward shaping as characterized in Formula~\eqref{eqn:PBRS}. 
\end{enumerate}

For training an agent, we use a simple fully connected feed-forward neural network with $21$ inputs, $2$ hidden layers with~$256$ neurons each, and~$5$ output neurons representing~$5$ discrete actions. This FNN model is trained with the Adam~\cite{kingma2014adam} optimization function in the Pytorch.
Lastly, we also use different $\gamma$ values (where $\gamma \in \{0.25,0.5,0.75,1\}$) for training to observe the effect of our PBRS reward shaping. 
In implementing Algorithm~\ref{algo:Framework}, we have used the following parameters: $\sig{timesteps}=200000$, $|B| = 5000$,   $\sig{uim}=400$, $\sig{uic}=400$, $\alpha=0.9$, $\sig{batch-size}= 200$. For Formula~\ref{kappa}, we set both $\kappa_1$ and $\kappa_2$ to be~$1$.

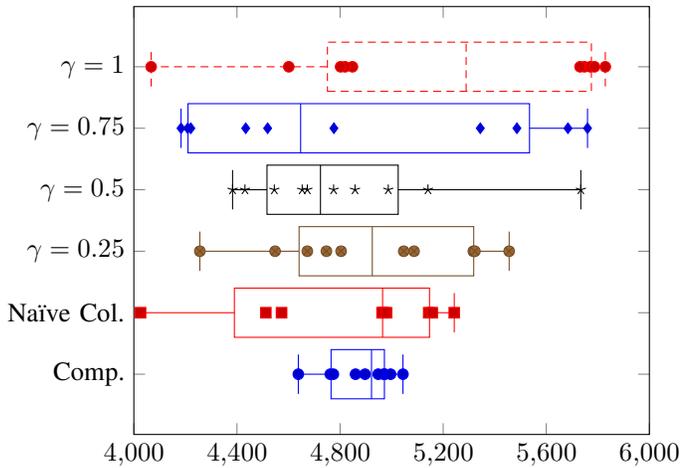
\begin{figure}
    \centering
    \begin{tikzpicture}
  \begin{axis}
    [
    xmin=4000,
    xmax=6000,
    xtick={4000,4400,4800,5200,5600,6000},
    ytick={1,2,3,4,5,6,7},
    yticklabels={Comp., Naïve Col., $\gamma=0.25$, 
    $\gamma=0.5$, $\gamma=0.75$, $\gamma=1$},
    ]
    \addplot+[
    boxplot prepared={
      median=4922,
      upper quartile=4972,
      lower quartile=4765,
      upper whisker=5044,
      lower whisker=4638
    },
    ] table[row sep=\\,y index=0] {4762\\ 4974\\ 4638\\ 4897\\ 5044\\4969\\4860\\4774\\4948\\4996\\};
    \addplot+[
    boxplot prepared={
      median=4965,
      upper quartile=5147,
      lower quartile=4390,
      upper whisker=5243,
      lower whisker=3326
    },
    ] table[row sep=\\,y index=0] {4967\\4963\\3326\\5144\\4026\\5243\\5158\\4980\\4573\\4512\\};
    \addplot+[
    boxplot prepared={
      median=4925,
      upper quartile=5318.5,
      lower quartile=4641,
      upper whisker=5456,
      lower whisker=4256
    },
    ] table[row sep=\\,y index=0] {5317\\4673\\4747\\4548\\5087\\5047\\4256\\5456\\4804\\5323\\};
    \addplot+[
    boxplot prepared={
      median=4724,
      upper quartile=5025.5,
      lower quartile=4516.5,
      upper whisker=5734,
      lower whisker=4383
    },
    ] table[row sep=\\,y index=0] {4987\\5141\\4673\\4431\\4653\\4383\\4858\\4775\\5734\\4545\\};
    \addplot+[
    boxplot prepared={
      median=4647,
      upper quartile=5535,
      lower quartile=4210,
      upper whisker=5760,
      lower whisker=4183
    },
    ] table[row sep=\\,y index=0] {4209\\5760\\4776\\4221\\4519\\5486\\5684\\4434\\5344\\4183\\};
    \addplot+[
    boxplot prepared={
      median=5289,
      upper quartile=5775,
      lower quartile=4751,
      upper whisker=5829,
      lower whisker=4067
    },
    ] table[row sep=\\,y index=0] {5748\\4848\\5829\\5771\\4067\\4819\\5731\\4802\\4601\\5787\\};
  \end{axis}
\end{tikzpicture}
\vspace{-3mm}
    \caption{Comparing the rewards created from different test case generation strategies}
    \label{fig:ploz}
    \vspace{-5mm}
\end{figure}

In our experiment, we create 10000 random initial states for simulation. We thereafter summarize the cumulative global reward of these randomly initialized experiments. 
The result of the evaluation is demonstrated in Figure~\ref{fig:ploz}, where the value of the $x$ coordinate reflects the sum of all simulation results reflected as $r(s,a)$ for each end state. Recall that we set both $\kappa_1$ and $\kappa_2$ to be~$1$, so the value essentially reflects $x - 2y$, where $x$ is the number of collisions with the AV, and $y$ is the occurrence of undesired collisions where two controlled test vehicles collide without AV involved. This means that the higher the number, the more effective the test case generation strategy. We can observe a mild win in cases where our proposed potential is integrated (with $\gamma = 1$) in the reward shaping scheme, where a large portion of the generated agents lead to overall better performance (around~$5800$) with the rest agents having performance to close to the competitive rewards. In addition, using our potential-based reward shaping scheme also largely demonstrates superiority over the  naïve collaborative method. However, one notable limitation is that in our very preliminary evaluation, the performance of the MARL agents can still have some variations, implying huge potential in improving the underlying training scheme.

\section{Conclusion and Future Work}\label{sec:concluding.remarks}
In this paper, we studied how reinforcement learning can be used to create test cases for autonomous driving. We adapted the principle of the multi-agent reinforcement learning paradigm and proposed a potential-based credit assignment method inspired by counterfactual analysis. Altogether it allows efficient training of agents to be used in test case generation. Our preliminary result demonstrated that in contrast to a fully competitive or naïve collaborative setup, our application-independent potential-based credit assignment leads to better performance in generating challenging scenarios. 

This work opens a new direction in generating test cases for autonomous driving. For follow-up work, the immediate extension is to test the methodology beyond simple AI Gym-like traffic simulators with realistic autonomous driving stacks and simulators such as Baidu Apollo and SVL. Another direction is to refine the credit assignment method for better performance. Yet another direction is to extend the framework with a multi-objective setup, allowing us to characterize different undesired situations ranging from collision to unintended acceleration.

\bibliographystyle{IEEEtran}

\end{document}